\documentclass{article} 
\usepackage{iclr2022_conference,times}

\usepackage{amsmath,amsfonts,bm}

\def\eqref#1{equation~\ref{#1}}

\def\1{\bm{1}}

\def\eps{{\epsilon}}

\DeclareMathAlphabet{\mathsfit}{\encodingdefault}{\sfdefault}{m}{sl}
\SetMathAlphabet{\mathsfit}{bold}{\encodingdefault}{\sfdefault}{bx}{n}

\DeclareMathOperator*{\argmin}{arg\,min}

\usepackage{epsfig}
\usepackage{graphicx}
\usepackage{amsmath}
\usepackage{amssymb}
\usepackage{ulem}

\usepackage{xcolor}
\usepackage{tabulary}
\usepackage{multirow}
\usepackage[ruled,vlined]{algorithm2e}

\usepackage{caption}
\usepackage{subcaption}
\usepackage{booktabs}

\newcommand{\ie}{\textit{i}.\textit{e}. }
\newcommand{\eg}{\textit{e}.\textit{g}. }
\definecolor{demphcolor}{RGB}{144,144,144}

\usepackage[pagebackref=true,breaklinks=true,colorlinks,bookmarks=false]{hyperref}

\title{Pseudo-Labeled Auto-Curriculum Learning for Semi-Supervised Keypoint Localization}

\author{
\centerline{Can Wang$^{1}$ \quad Sheng Jin$^{2,1}$ \quad Yingda Guan$^{1}$ \quad Wentao Liu$^{1}$\thanks{Corresponding author.} \quad Chen Qian$^{1}$} \\
\centerline{\textbf{Ping Luo$^{2}$ \quad Wanli Ouyang$^{3}$}} 
\\
\centerline{$^{1}$SenseTime Research and Tetras.AI \quad 
$^{2}$The University of Hong Kong \quad $^{3}$The University of Sydney}
\\
\centerline{\texttt{\{wangcan, jinsheng\}@tetras.ai} \quad \texttt{\{guanyingda, liuwentao, qianchen\}@sensetime.com}} \\
\centerline{ \texttt{pluo@cs.hku.hk} \quad \texttt{wanli.ouyang@sydney.edu.au} }
}

\iclrfinalcopy 
\begin{document}

\maketitle

\begin{abstract}

Localizing keypoints of an object is a basic visual problem. However, supervised learning of a keypoint localization network often requires a large amount of data, which is expensive and time-consuming to obtain. To remedy this, there is an ever-growing interest in semi-supervised learning (SSL), which leverages a small set of labeled data along with a large set of unlabeled data. Among these SSL approaches, pseudo-labeling (PL) is one of the most popular. PL approaches apply pseudo-labels to unlabeled data, and then train the model with a combination of the labeled and pseudo-labeled data iteratively. 
The key to the success of PL is the selection of high-quality pseudo-labeled samples. Previous works mostly select training samples by manually setting a single confidence threshold. We propose to automatically select reliable pseudo-labeled samples with a series of dynamic thresholds, which constitutes a learning curriculum.
Extensive experiments on six keypoint localization benchmark datasets demonstrate that the proposed approach significantly outperforms the previous state-of-the-art SSL approaches.

\end{abstract}


\section{Introduction}

Keypoints (also termed as landmarks) are a popular representation of objects that precisely represent locations of object parts and contain concise information about shapes and poses. Example keypoints are "right shoulder" on a human body or the "tail tip" of a cat. Keypoint localization is the basis of many visual tasks, including action recognition~\citep{yan2018spatial}, fine-grained classification~\citep{gavves2013fine,gavves2015local}, pose tracking~\citep{jin2017towards,jin2019multi} and re-identification~\citep{zhao2017spindle}.

Keypoint localization has achieved great success with the advent of deep learning in recent years~\citep{newell2016stacked,xiao2018simple,duan2019trb,sun2019deep,jin2020differentiable,xu2021vipnas,geng2021bottom,li2021human}. However, the success of deep networks relies on vast amounts of labeled data, which is often expensive and time-consuming to collect. Semi-supervised learning (SSL) is one of the most important approaches for solving this problem. It leverages extensive amounts of unlabeled data in addition to sparsely labeled data to obtain gains in performance. Pseudo-labeling (PL) has become one of the most popular SSL approaches due to its simplicity. PL-based methods iteratively add unlabeled samples into the training data by pseudo-labeling them with a model trained on a combination of labeled and pseudo-labeled samples. 

PL-based methods commonly require a predefined handpicked threshold~\citep{lee2013pseudo, oliver2018realistic}, to filter out low-confidence noisy predictions. However, a single fixed threshold does not take into account the dynamic capacity of the current model for handling noisy pseudo-labels, leading to sub-optimal performance.
In this work, we borrow ideas from Curriculum Learning (CL)~\citep{bengio2009curriculum} and design our curriculum as a series of thresholds for PL, which is tuned according to the feedback from the model. CL is a widely used strategy to control the model training pace by selecting from easier to harder samples. With a carefully designed curriculum, noticeable improvement is obtained. 
However, traditional CL methods suffer from hand-designed curricula, which heavily rely on expertise and detailed analysis for speciﬁc domains. Manual curriculum design based on handcrafted criteria is always tedious and sub-optimal.
Moreover, curriculum design (or threshold setting) is complicated. 
High-confidence pseudo-labels typically correspond to easier samples with clean labels, while low-confidence pseudo-labels correspond to harder samples with noisy labels. How to design a curriculum to balance the correctness, representativeness, and difficulty of pseudo-labeled data is an open problem. 
This paper is devoted to tackling the aforementioned problem, \ie how to automatically learn an optimal learning curriculum for pseudo-labeling in a data-driven way. To this end, we propose a novel method, called Pseudo-Labeled Auto-Curriculum Learning (PLACL). PLACL formulates the curriculum design problem as a decision-making problem and leverages the reinforcement learning (RL) framework to solve it. 

Additionally, PL-based methods suffer from confirmation bias~\citep{tarvainen2017mean}, also known as noise accumulation~\citep{zhang2016enhanced}, and concept drift~\citep{cascante2020curriculum}. This long-standing issue stems from the use of noisy or incorrect pseudo-labels in subsequent training stages. As a consequence, the noise accumulates and the performance degrades as the learning process evolves over time. To mitigate this problem, we propose the cross-training strategy which alternatively performs pseudo-label prediction and model training on separate sub-datasets.

We benchmark PLACL on six keypoint localization datasets, including  LSPET~\citep{lsp-extended}, MPII~\citep{mpii}, CUB-200-2011~\citep{cub-200-2011}, ATRW~\citep{ATRW}, MS-COCO~\citep{coco}, and AnimalPose~\citep{cao2019cross}. We empirically show that PLACL is general and can be applied to various keypoint localization tasks (human and animal pose estimation) and different keypoint localization networks. With a simple yet effective search paradigm, our method significantly boosts the keypoint estimation performance and achieves superior performance to other SSL methods.
We hope our method will inspire the community to rethink the potential of PL-based methods for semi-supervised keypoint localization.

Our main contributions can be summarized as follows:
\begin{itemize}

\item We propose Pseudo-Labeled Auto-Curriculum Learning (PLACL). It is an an automatic pseudo-labeled data selection method, which learns a series of dynamic thresholds (or curriculum) via reinforcement learning. To the best of our knowledge, this is the ﬁrst work that explores automatic curriculum learning for semi-supervised keypoint localization.

\item We propose the cross-training strategy for pseudo-labeling to mitigate the long-standing problem of confirmation bias.

\item Extensive experiments on a wide range of popular datasets demonstrate the superiority of PLACL over the previous state-of-the-art SSL approaches. In addition, PLACL is model-agnostic and can be easily applied to different keypoint localization networks. 
\end{itemize}

\section{Related Works}

\subsection{Semi-supervised keypoint localization}
Keypoint localization focuses on predicting the keypoints of detected objects, \eg human body parts~\citep{li2019crowdpose, jin2020whole}, facial landmarks~\citep{bulat2017far}, hand keypoints~\citep{zimmermann2017learning} and animal poses~\citep{cao2019cross}. However, training a keypoint localization model often requires a large amount of data, which is expensive and time-consuming to collect. Semi-supervised keypoint localization is one of the most promising ways to solve this problem. Semi-supervised keypoint localization can be categorized into consistency regularization based methods and pseudo-labeling based methods. Consistency regularization methods~\citep{honari2018improving,moskvyak2020semi} assume that the output of the model should not be invariant to realistic perturbations. These approaches typically rely on modality-specific augmentation techniques for regularization. Pseudo-labeling methods~\citep{ukita2018semi,dong2019teacher,cao2019cross,li2021synthetic} use labeled data to predict the labels of the unlabeled data, and then train the model in a supervised way with a combination of labeled and selected pseudo-labeled data. Our approach also builds upon pseudo-labeling methods. In contrast to previous works, we propose to learn pseudo-labeled data selection via reinforcement learning.

\subsection{Curriculum Learning}
Curriculum learning is firstly introduced by~\cite{bengio2009curriculum}. It is a training strategy that trains machine learning models from easy to complex samples, imitating human education. The curriculum is often pre-determined by heuristics~\citep{khan2011humans,bengio2009curriculum,spitkovsky2009baby}. However, it requires expert domain knowledge and exhaustive trials to find a good curriculum suitable for a specific task and its dataset. Recently, automatic curriculum learning methods are introduced to break through these limits. Popular ones include self-paced learning methods~\citep{kumar2010self,jiang2014easy,zhao2015self} and reinforcement learning (RL) based methods~\citep{graves2017automated,matiisen2019teacher,fan2018learning}. Our approach can be categorized as RL-based methods. Unlike previous works that focus on supervised learning, our approach is designed for the SSL paradigm. Our work is mostly related to Curriculum Labeling~\citep{cascante2020curriculum}. It adopts a \textit{hand-crafted} curriculum based on Extreme Value Theory (EVT) to facilitate model training. In contrast, we propose an automatic curriculum learning approach by searching for dynamic thresholds for pseudo-labeling. In addition, the curriculum of~\citep{cascante2020curriculum} is coarse-grained on the round level, while our curriculum is fine-grained on the epoch level.

\subsection{Reinforcement learning for AutoML}
Reinforcement learning (RL) has shown impressive results in a range of applications. Well-known examples include game playing \citep{mnih2015human, silver2016mastering, silver2017mastering} and robotics control\citep{schulman2015trust, lillicrap2015continuous}. Recent works have employed RL to the AutoML, automating the design of a machine learning system, \eg searching for neural architectures~\citep{zoph2016neural, zoph2018learning, baker2016designing, pham2018efficient}, augmentation policies~\citep{cubuk2019autoaugment}, activation functions~\citep{ramachandran2017searching}, loss functions~\citep{li2019lfs,li2020auto}, and training hyperparameters~\citep{dong2020autohas}. In contrast to these works, we apply RL to the automatic selection of pseudo-labeled data in the context of pseudo-labeling.

\section{Pseudo-labeled Auto-Curriculum Learning (PLACL)}

\subsection{Overview}

\begin{figure}[t]
    \begin{center}
        \includegraphics[width=\linewidth]{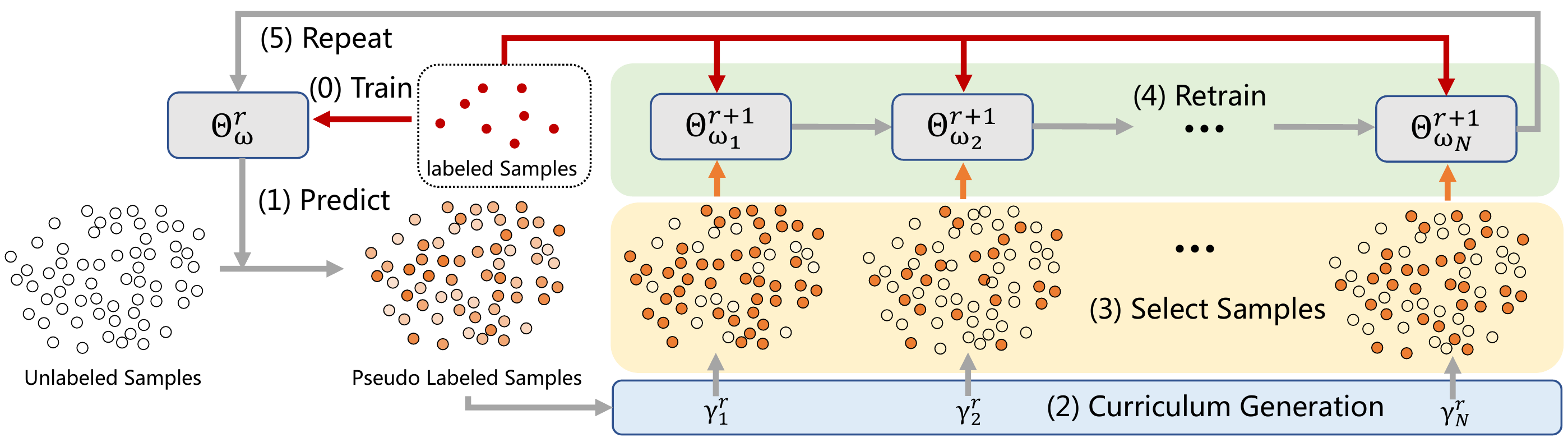}
    \end{center}
    \caption{Pseudo-Labeled Auto-Curriculum Learning (PLACL). (0) In the initial round, the model $\Theta_\omega^0$ is pre-trained on the labeled data. And for the $r$-th round, (1) the trained network $\Theta_\omega^r$ is used to predict pseudo-labels for unlabeled data. (2) A learning curriculum consisting of a series of thresholds ($\gamma_i^r$) is generated. (3) Reliable pseudo-labeled data is selected by the searched curriculum. (4) A new model $\Theta_\omega^{r+1}$ is retrained using both the labeled samples and selected pseudo-labeled samples. (5) This process is repeated by re-labeling unlabeled data using the new model.}
    \label{fig:framework}
\end{figure}

Our PLACL algorithm is illustrated in Fig.~\ref{fig:framework}. The training process consists of $R$ self-training rounds and each round consists of $N$ training epochs. (0) In the initial round ($r=0$), we pre-train a keypoint localization network $\Theta_\omega^0$ on the labeled data, where $\omega$ denotes the weights of the network. And for the $r$-th round, (1) The trained network $\Theta_\omega^r$ is used to predict pseudo-labels for unlabeled data. (2) We adopt reinforcement learning (RL) to automatically generate the learning curriculum. Specifically, our curriculum ($\Gamma^r$) consists of a series of thresholds for pseudo-labeled data selection. $\Gamma^r = [\gamma_1^r, \dots, \gamma_{N}^r]$, where $\gamma_{i}^r \in [0, 1]$ is the threshold for each epoch $i$. (3) We then select reliable pseudo-labeled data by the searched curriculum. (4) We retrain a new model ($\Theta_\omega^{r+1}$) using both the labeled samples and selected pseudo-labeled samples. (5) This process is repeated for $R$ rounds.

\subsection{Pseudo-label Selection for semi-supervised keypoint localization}
\label{sec:sskl}

We denote the labeled dataset with $N_l$ samples as $\mathbb{D}_{l}=\left\{\left.\left(\boldsymbol{I}_{i}^{l}, \boldsymbol{Y}_{i}^{l}\right)\right|_{i=1} ^{N_{l}}\right\}$,where $\boldsymbol{I}_{i}$  and $\boldsymbol{Y}_{i}$ denote the $i$-th training image and its keypoint annotations (the x-y coordinates of $K$ keypoints). The $N_u$ unlabeled images are denoted as $\mathbb{D}_{u}=\left\{\left.\left(\boldsymbol{I}_{i}^{u}\right)\right|_{i=1} ^{N_{u}}\right\}$,  which are not associated with any ground-truth keypoint labels. Generally, we have $\left|N_{l}\right| \ll\left|N_u\right|$.

Pseudo-labeling based method builds upon the general idea of self-training~\citep{mclachlan1975iterative}, where the keypoint localization network $\Theta_\omega$ goes through multiple rounds of training. 
In the initialization round, the model is first trained with the small labeled training set $\mathbb{D}_{train}$ = $\mathbb{D}_{l}$ in a usual supervised manner. 
In subsequent rounds, the trained model is used to estimate labels for the unlabeled data $\tilde{\mathbb{D}}_{u} = \left\{\left.\left(\boldsymbol{I}_{i}^{u}, \tilde{\boldsymbol{Y}}_{i}^{u}\right)\right|_{i=1} ^{N_{u}}\right\}$. Here, we omit the superscript $r$ for simplicity.
Specifically, given an unlabeled image $\boldsymbol{I}_{i}^{u}$, the trained keypoint localization network $\Theta_\omega$ predicts $K$ heatmaps.
Each heatmap is a 2D Gaussian centered on the joint location, which represents the confidence of the $k$-th keypoint. The output pseudo-labeled keypoint location ($\tilde{\boldsymbol{Y}}_{i}^{u}$) is the highest response in the heatmap space. And the confidence score $\mathcal{C}\left(\Theta_{\omega} (\boldsymbol{I}_{i}^{u}) \right)$ is the response value at the keypoint location.

Then, pseudo-label selection process is adopted. Let $\boldsymbol{g}=\left[g_1, ..., g_{N_u}\right] \subseteq \{0,1\}^{N_u}$ be a binary vector representing the selection of pseudo-labels, where $g_{i}$ denotes whether the keypoint prediction on $\boldsymbol{I}_{i}^{u}$ is selected.

\begin{equation}
g_{i} =\left\{\begin{array}{l}1, \text { if } \mathcal{C}\left(\Theta_\omega (\boldsymbol{I}_{i}^{u}) \right)> \gamma \\ 0, \text { otherwise }\end{array}\right.
\label{eq:selection_criterion}
\end{equation}

where $\gamma \in \left( 0, 1 \right)$ is the confidence threshold. Pseudo-labeled samples with higher confidence are added to the training set. 

\begin{equation}
    \mathbb{D}_{train}=\left\{\left.\left(\boldsymbol{I}_{i}^{l}, \boldsymbol{Y}_{i}^{l}\right)\right|_{i=1} ^{N_{l}}\right\} \cup \left\{\left.\left(\boldsymbol{I}_{i}^{u}, \tilde{\boldsymbol{Y}}_{i}^{u}\right)\right|_{i=1} ^{N_{u}} \text{ where } g_i = 1  \right\}.
    \label{eq:dataset_train}
\end{equation}

Then the keypoint localization network is retrained with a combination of labeled and pseudo-labeled training data $\mathbb{D}_{train}$.

\subsection{Cross-training strategy} 
\label{sec:cross-training}
In this section, we introduce the cross-training strategy in our curriculum learning framework. 
In a typical self-training round, the model predicts noisy pseudo-labels which are used in subsequent training stages. Since the pseudo-label prediction is performed on a dataset of \textit{known data} (on which training is performed), the noise accumulates in a positive feedback loop. This causes the long-standing issue of confirmation bias~\citep{tarvainen2017mean}. To mitigate this problem, we propose the cross-training strategy. Specifically, we randomly partition the unlabeled data $\mathbb{D}_{u}$ into two complementary subsets $\mathbb{D}_{u} ^{(1)}$ and $\mathbb{D}_{u} ^{(2)}$. The partition is fixed across all the training rounds. The self-training conducts alternatively for $\mathbb{D}_{u} ^{(1)}$ and $\mathbb{D}_{u} ^{(2)}$. For $r = \{1, \dots, R\}$, when $r$ is odd, we use a combination of $\tilde{\mathbb{D}}_{u} ^{(1)}$ and $\mathbb{D}_{l}$ to train the model, and the trained model predicts pseudo-labels for the next round on $\mathbb{D}_{u} ^{(2)}$. When $r$ is even, we use a combination of $\tilde{\mathbb{D}}_{u} ^{(2)}$ and $\mathbb{D}_{l}$ to train the model, and perform pseudo-label prediction on $\mathbb{D}_{u} ^{(1)}$. 
Before each round, the model parameters are re-initialized with random weights following~\cite{cascante2020curriculum} to avoid noise accumulation.

\subsection{Curriculum Residual Learning}
Directly learning for $R$ rounds of curriculum parameters separately can be inefficient. Considering that the model is reinitialized and retrained in each self-training round, we assume that the optimal curricula for different rounds should have some similar patterns. We propose a greedy multi-step searching algorithm. For round $r$, we use the searched curriculum in the previous round $r-1$ as the base curriculum to guide the searching of current curricula $\Gamma^r$. Inspired by ResNet~\citep{he2016deep}, we propose the curriculum residual learning strategy. Formally, we learn a bias term ${\Delta \Gamma}^r$ around the base curriculum, \ie $\Gamma^r = (\Gamma^{r-1})^* +  {\Delta \Gamma}^r$, where $(\Gamma^{r-1})^*$ means the searched optimal curriculum for round $r-1$, and $(\Gamma^0)^*$ is initialized with all zeros. We empirically find that this strategy accelerates the model convergence speed and achieves marginally better performance. To further reduce the search space, we choose every $G$ epochs as an epoch group, which shares the same threshold parameters for pseudo-labeled data selection. In total, we have $N_G$ epoch groups, and the size of each epoch group is $G$. Therefore, the search space is reduced by a factor of $G$, where $G=10$ in our implementation. 

\subsection{Curriculum Search via Reinforcement Learning}

Our PLACL can be formulated as an optimization problem shown in Eq.~\ref{eq:PLACL}. In the inner-loop, we optimize the weights $\omega$ of the keypoint localization network $\Theta_\omega$ to minimize the training loss L (see Alg.~\ref{alg:inner}). 
In the outer-loop, we apply proximal policy optimization (PPO2) algorithm~\citep{schulman2017proximal} to search for the curriculum $\Gamma$ that maximize the evaluation metric $\xi$ (\eg PCK) on the validation set $\mathbb{D}_\text{val}$ (see Alg.~\ref{alg:PLACL}).

\begin{equation} \label{eq:PLACL}
\begin{aligned}
    \max_\Gamma~ \xi(\Gamma) =~ \xi(\Theta_{\omega^*(\Gamma)}; \mathbb{D}_\text{val}), \quad
    \text{s.t.} \quad \omega^*(\Gamma) =~ \argmin_\omega~ L(\Theta_\omega; \mathbb{D}_{\text{train}}, \Gamma).
\end{aligned}
\end{equation}

The training consists of $R$ rounds. In each round $r$, the PPO2 search process consists of $T$ sampling steps. In each step, $M$ sets of parameters are sampled independently from a truncated normal distribution~\citep{nakano2012control, fujita2018clipped}, $\Delta \Gamma^r \sim \mathcal{N}_{\operatorname{trunc}[0,1]}\left(\mu_{t}^r, \sigma^{2} I\right)$, where $\mu_{t}^r$ and $\sigma^{2} I$ are the mean and covariance ($\sigma$ is fixed to $0.2$ in practice). 
These sampled parameters are used to construct $M$ different training curricula for training $M$ keypoint localization networks separately. Then the mean of the distribution is updated by PPO2 algorithm according to the evaluation score of the $M$ networks.

The objective function of PPO2 is formulated in Eq.~\ref{eq:objective-function}. 
\begin{equation}
\begin{aligned}
J\left(\mu^r \right) =\mathbb{E}_{\pi}\left[ \min \left(\frac{ \pi_{\mu^r} \left(\Delta \Gamma_{j}^r ; \mu^r, \sigma^{2} I\right)}{\pi_{\mu_t^r}\left(\Delta \Gamma_{j}^r ; \mu_{t}^r, \sigma^{2} I\right)} \tilde{\xi} \left(\Gamma_{j}^r\right), \text { CLIP }\left(\frac{\pi_{\mu^r}\left(\Delta \Gamma_{j}^r ; \mu^r, \sigma^{2} I\right)}{\pi_{\mu_t^r}\left(\Delta \Gamma_{j}^r ; \mu_{t}^r, \sigma^{2} I\right)}, 1-\epsilon, 1+\epsilon\right) \tilde{\xi}\left(\Gamma_{j}^r\right) \right)\right],
\end{aligned}
\label{eq:objective-function}
\end{equation}

where the function $\operatorname{CLIP}(x, 1-\epsilon, 1+\epsilon)$ clips $x$ to be no more than $1+ \epsilon$ and no less than $1-\epsilon$. Following the common practice~\citep{li2020auto}, the mean reward is subtracted for better convergence. $\tilde{\xi}\left( \Gamma_{j}^r\right) = \xi \left( \Gamma_{j}^r\right) -  \frac{1}{M}\sum_{j=1}^{M}{\xi}\left(\Gamma_{j}^r\right)$ and the policy $\pi_{\mu^r}$ is defined as the probability density function (PDF) of the truncated normal distribution. PPO2 enforces the \textit{probability ratio} between old and new policies $\pi_{\mu^r}\left(\Delta \Gamma_{j}^r ; \mu^r, \sigma^{2} I\right) / \pi_{\mu_t^r}\left(\Delta \Gamma_{j}^r ; \mu_t^r, \sigma^{2} I\right)$ to stay within a small interval to control the size of each policy update. We then compute the gradients and update the parameters by $\mu_{t+1}^r \leftarrow \mu_t^r +\alpha \nabla_{\mu^r} J\left(\mu^r\right)$ with a learning rate of $\alpha > 0$. 
After $T$ sampling steps, we choose $\mu_t^r$ with the highest average evaluation score as $(\Gamma^r)^{*}$. And $(\Gamma^r)^{*}$ is used as the base curriculum for the next round.
And after $R$ rounds, our final optimal curriculum is obtained, $\Gamma^{*}= [(\Gamma^{1})^*, \dots, (\Gamma^{R})^*]$.

\paragraph{Training details}
In the training phase, the keypoint localization network and the curriculum search policy are simultaneously optimized. For the outer-loop, the PPO2~\citep{schulman2017proximal} search procedure is conducted for $T = 16$ sampling steps, and in each step $M = 8$ sets of parameters (curriculum)  are sampled. The clipping threshold is $\eps = 0.2$, and $\mu_{t+1}^{r}$ is updated with the learning rate of $\alpha = 0.2$. We empirically use $R=6$ self-training rounds, and group size $G=10$ for curriculum search. For the inner-loop, we follow the common practice~\citep{sun2019deep,mmpose2020} to train the keypoint localization network with Mean-Squared Error (MSE) loss for $N=210$ epochs per round. Adam~\citep{adam} with a learning rate of $0.001$ is adopted. We reduce the learning rate by a factor of $10$ at the 170-th and 200-th epochs. 
Although the RL search process increases the training complexity, the total training cost is not too high (only 1.5 days with 32 NVIDIA Tesla V100 GPUs).
More detailed training settings for each task are provided in \S\ref{sec:exp_tabel1} and \S\ref{sec:exp_animalpose}.

\section{Experiments}

\begin{table*}[tb]
\caption{Keypoint localization with different percentage of labeled images. We report mean and standard deviation from three runs for different randomly sampled labeled subsets. Pseudo-labeling (PL) based methods are not evaluated for 100\% of labeled data because there is no unlabeled data to generate pseudo-labels for. The results marked with `*' are from~\citep{moskvyak2020semi}.}
\label{tab:all_results}
\begin{center} 
 \resizebox{1.\textwidth}{!}{
\begin{tabular}{llllll} 
 & \multicolumn{5}{c}{Percentage of labeled images} \\ 
 Method & \multicolumn{1}{c}{5\%} & \multicolumn{1}{c}{10\%} & \multicolumn{1}{c}{20\%} & \multicolumn{1}{c}{50\%} & \multicolumn{1}{c}{100\%} \\ 

\toprule 
\multicolumn{6}{c}{\textbf{Dataset 1: LSPET } } \\
HRNet$^{*}$ \citep{sun2019deep}&40.19$\pm$1.46 &45.17$\pm$1.15 &55.22$\pm$1.41 &62.61$\pm$1.25 &72.12$\pm$0.30 \\ 
ELT$^{*}$ \citep{honari2018improving}&41.77$\pm$1.56 &47.22$\pm$0.91 &57.34$\pm$0.94 &66.81$\pm$0.62 &72.22$\pm$0.13 \\ 
Gen$^{*}$ \citep{jakab2018unsupervised}&61.01$\pm$1.41 &67.75$\pm$1.00 &68.80$\pm$0.91 &69.70$\pm$0.77 &72.25$\pm$0.55 \\ 
SSKL$^{*}$~\citep{moskvyak2020semi} &66.98$\pm$0.94 &69.56$\pm$0.66 &71.85$\pm$0.33 &72.59$\pm$0.56 &74.29$\pm$0.21 \\
PL$^{*}$ \citep{radosavovic2018data} & 37.36$\pm$1.89 & 42.05$\pm$1.68 & 48.86$\pm$1.23 & 64.45$\pm$0.96 & - \\
CL \citep{cascante2020curriculum} & 61.27$\pm$1.54  & 65.43$\pm$1.19  & 69.14$\pm$0.93 &   70.29$\pm$1.18 & - \\ 
PLACL (Ours) & \textbf{70.76$\pm$1.47}  & \textbf{71.91$\pm$1.15}  & \textbf{72.30$\pm$0.88}  & \textbf{72.73$\pm$1.23}  & - \\ 

\midrule 
\multicolumn{6}{c}{\textbf{Dataset 2: MPII } } \\
HRNet$^{*}$ \citep{sun2019deep}&66.22$\pm$1.60 &69.18$\pm$1.03 &71.83$\pm$0.87 &75.73$\pm$0.35 &81.11$\pm$0.15 \\ 
ELT$^{*}$ \citep{honari2018improving}&68.27$\pm$0.64 &71.03$\pm$0.46 &72.37$\pm$0.58 &77.75$\pm$0.31 &81.01$\pm$0.15 \\
Gen$^{*}$ \citep{jakab2018unsupervised}&71.59$\pm$1.12 &72.63$\pm$0.62 &74.95$\pm$0.32 &79.86$\pm$0.19 &80.92$\pm$0.32 \\ 
SSKL$^{*}$~\citep{moskvyak2020semi} &74.15$\pm$0.83 &76.56$\pm$0.48 &78.46$\pm$0.36 &80.75$\pm$0.32 &82.12$\pm$0.14 \\
PL$^{*}$ \citep{radosavovic2018data} & 62.44$\pm$1.75 & 64.78$\pm$1.44 & 69.35$\pm$1.11 & 77.43$\pm$0.48 & - \\ 
CL \citep{cascante2020curriculum} & 72.03$\pm$1.56  & 73.15$\pm$0.95  & 75.80$\pm$0.92 &  77.49$\pm$0.35 & - \\ 
PLACL (Ours) & \textbf{77.83$\pm$1.41}  & \textbf{78.36$\pm$0.92}  & \textbf{79.68$\pm$0.72}  & \textbf{80.81$\pm$0.24}  & - \\

\midrule 
\multicolumn{6}{c}{\textbf{Dataset 3: CUB-200-2011 } } \\
HRNet$^{*}$ \citep{sun2019deep}&85.77$\pm$0.38 &88.62$\pm$0.14 &90.18$\pm$0.22 &92.60$\pm$0.28 &93.62$\pm$0.13 \\
ELT$^{*}$ \citep{honari2018improving}&86.54$\pm$0.34 &89.48$\pm$0.25 &90.86$\pm$0.13 &92.26$\pm$0.06 &93.77$\pm$0.18 \\ 
Gen$^{*}$ \citep{jakab2018unsupervised}&88.37$\pm$0.40 &90.38$\pm$0.22 &91.31$\pm$0.21 &92.79$\pm$0.14 &93.62$\pm$0.25 \\ 
SSKL$^{*}$~\citep{moskvyak2020semi} &91.11$\pm$0.33 &91.47$\pm$0.36 &92.36$\pm$0.30 &92.80$\pm$0.24 &93.81 $\pm$0.13 \\
PL$^{*}$ \citep{radosavovic2018data} & 86.31$\pm$0.45 & 89.51$\pm$0.32 & 90.88$\pm$0.28 & 92.78$\pm$0.27 & - \\
CL \citep{cascante2020curriculum} & 91.46$\pm$0.41  & 92.35$\pm$0.34  & 92.74$\pm$0.27 & 92.97$\pm$0.21  & - \\ 
PLACL (Ours) & \textbf{93.01$\pm$0.33} & \textbf{93.28$\pm$0.29} & \textbf{93.45$\pm$0.25} & \textbf{93.84$\pm$0.18} & - \\

\midrule 
\multicolumn{6}{c}{\textbf{Dataset 4: ATRW } } \\
HRNet$^{*}$ \citep{sun2019deep}&69.22$\pm$0.87 &77.55$\pm$0.84 &86.41$\pm$0.45 &92.17$\pm$0.18 &94.44$\pm$0.10 \\ 
ELT$^{*}$ \citep{honari2018improving}&74.53$\pm$1.24 &80.35$\pm$0.96 &87.98$\pm$0.47 &92.80$\pm$0.21 &94.75$\pm$0.14 \\ 
Gen$^{*}$ \citep{jakab2018unsupervised}&89.54$\pm$0.57 &90.48$\pm$0.49 &91.16$\pm$0.13 &92.27$\pm$0.24 &94.80$\pm$0.13 \\ 
SSKL$^{*}$~\citep{moskvyak2020semi} &92.57$\pm$0.64 &94.29$\pm$0.66 &94.49$\pm$0.36 &94.63$\pm$0.18 &95.31$\pm$0.12 \\
PL$^{*}$ \citep{radosavovic2018data} & 67.97$\pm$1.07 & 75.26$\pm$0.74 & 84.69$\pm$0.57 & 92.15$\pm$0.24 & - \\
CL \citep{cascante2020curriculum} & 87.01$\pm$1.08  &  89.13$\pm$0.94  & 92.34$\pm$0.51 & 93.57$\pm$0.26  & - \\ 
PLACL (Ours) & \textbf{94.37$\pm$0.86} & \textbf{94.59$\pm$0.80}  & \textbf{94.85$\pm$0.48}  & \textbf{95.01$\pm$0.17}  & - \\ 

\midrule 
\multicolumn{6}{c}{\textbf{Dataset 5: MS-COCO'2017 } } \\
HRNet \citep{sun2019deep}   & 62.44$\pm$1.26 & 66.02$\pm$1.07 & 69.62$\pm$0.84 & 72.81$\pm$0.73 & 74.61$\pm$0.58  \\ 
CL \citep{cascante2020curriculum} & 64.47$\pm$1.18  & 67.82$\pm$0.95 & 70.36$\pm$0.89 & 72.92$\pm$0.84 & \\ 
PLACL (Ours) & \textbf{69.39$\pm$1.03}  & \textbf{70.11$\pm$0.89}  & \textbf{71.84$\pm$0.66}  & \textbf{73.42$\pm$0.57}  & - \\ 

\bottomrule

\end{tabular} 
}
\end{center}
\end{table*}

\subsection{Datasets and Evaluation Metrics}

\paragraph{Datasets:} To show the versatility of PLACL, we conduct experiments on 5 diverse datasets. 
\textbf{LSPET} (Leeds Sports Pose Extended Dataset)~\citep{lsp-original,lsp-extended} consists of images of people doing sports activities. We use 10,000 images from~\citep{lsp-extended} for training and 2,000 images from~\citep{lsp-original} for validation and testing.
\textbf{MPII} Human Pose dataset~\citep{mpii} is a well-known benchmark for human pose estimation. The images are collected from YouTube videos, showing people doing daily human activities. We follow~\citep{moskvyak2020semi} to use 10,000 random images from MPII \texttt{train} for training, 3,311 images from MPII \texttt{train} for validation and MPII \texttt{val} for evaluation.
\textbf{CUB-200-2011} (Caltech-UCSD Birds-200-2011)~\citep{cub-200-2011} dataset is a well-known dataset for SSL. It consists of 200 fine-grained bird species with 15 keypoint annotations. We follow~\citep{moskvyak2020semi} to split dataset into training (100 categories with 5,864 images), validation (50 categories with 2,958 images) and testing (50 categories with 2,966 images). 
\textbf{ATRW}~\citep{ATRW} dataset contains images of 92 Amur tigers captured from multiple wild zoos in challenging and unconstrained conditions. For each tiger, 15 body keypoints are annotated. The dataset consists of 3,610 images for training, 516 for validation, and 1,033 for testing.
\textbf{MS-COCO'2017}~\citep{coco} is a popular large-scale benchmark for human pose estimation, which contains over 150,000 annotated people. We randomly select 500 images from COCO \texttt{train} for validation, the remaining training set (115k images) for training, and COCO \texttt{val} (5k images) for evaluation. We use this dataset to validate the applicability of our approach on large-scale data. 
\textbf{AnimalPose}~\citep{cao2019cross} dataset contains 5,517 instances of five animal categories: dog, cat, horse, sheep, and cow. It consists of 2,798 images for training, and the 810 images for validation and 1,000 images for testing. We use it to test the generalization and domain-transfer capacity of our proposed method.

\paragraph{Evaluation Metrics:}
\textbf{PCK} (Probability of Correct Keypoint): A detected keypoint is considered correct if the distance between the predicted and true keypoint is within a certain threshold ($\alpha l$), where $\alpha$ is a constant and $l$ is the longest side of the bounding box. We adopt PCK@0.1 ($\alpha$ = 0.1) for LSPET~\citep{lsp-extended}, CUB-200-2011~\citep{cub-200-2011}, and ATRW~\citep{ATRW} datasets.
\textbf{PCKh} is adapted from PCK, where $l$ is the head size that corresponds to 60\% of the diagonal length of the ground-truth head box. We adopt PCKh@0.5 ($\alpha$ = 0.5) for MPII~\citep{mpii} dataset.
Standard \textbf{AP} (Average Precision) is another commonly used evaluation metric. It is based on object keypoint similarity (OKS), which measures the distance between predicted keypoints and ground-truth keypoints normalized by the scale of the object. We use mAP for AnimalPose~\citep{cao2019cross} datasets.

\subsection{Comparisons with the State-of-the-Art SSL Approaches}\label{sec:exp_tabel1} 

In Table~\ref{tab:all_results}, we compare with the supervised baseline (HRNet~\citep{sun2019deep}) and other state-of-the-art SSL approaches. We experiment with different percentages of labeled images (5\%, 10\%, 20\%, 50\%, and 100\%). For fair comparisons, all results are obtained using HRNet-w32 backbone with the input size of $256 \times 256$. We follow~\citep{moskvyak2020semi} to prepare datasets and exclude half body transforms, and testing tricks (post-processing, and flip testing).

\textbf{Comparisons with consistency regularization methods.} 
ELT~\citep{honari2018improving} (equivariant landmark transformation) loss encourages the model to output keypoints that are equivariant to input transformations. Gen~\citep{jakab2018unsupervised} learns to extract geometry-related features through conditional image generation. SSKL~\citep{moskvyak2020semi} learns the pose invariant keypoint representations with semantic keypoint consistency constraints. These consistency regularization methods have shown superior results over the supervised baseline, however, they are inferior to our PLACL method on all datasets and different percentages of labeled samples. Especially we show that PLACL is mostly effective for low data regimes. For example, in CUB-200-2011 dataset, PLACL with only 5\% labeled data achieves better performance (93.01 vs 92.80) than SSKL with 50\% labeled data. And in LSPET dataset, we show that PLACL improves the performance of baseline by a large margin from 40.19 to 70.76 with 5\% labeled images.

\textbf{Comparisons with pseudo-labeling method.} 
We also compare with a pseudo-labeling (PL) baseline~\citep{radosavovic2018data}. Overall PLACL significantly outperforms the PL baseline on all datasets. As pointed out by~\cite{moskvyak2020semi}, the vanilla PL approach does not perform well for the keypoint localization task with a low data regime, due to the lack of an effective pseudo-label selection scheme.
Instead, PLACL is able to automatically select high-quality pseudo-labeled samples, which is the key to the success of pseudo-labeling based methods. 

\textbf{Comparisons with curriculum-learning method.} 
Curriculum Labeling (CL)~\citep{cascante2020curriculum} is a recently proposed approach that applies a \textit{hand-crafted} curriculum to facilitate training of SSL. We observe that our proposed PLACL significantly outperforms CL, which validates the effectiveness of our proposed \textit{automatic} curriculum learning. 

\textbf{Experiments on large-scale datasets.}
Inspired by~\cite{zhou2020econas}, in order to decrease the RL curriculum search cost for the large-scale MS-COCO~\citep{coco} and full MPII~\citep{mpii} datasets, we use a light proxy task with reduced number of training samples (5k) for RL curriculum search. After the search procedure, we re-train the keypoint networks with the searched curriculum on the full training set and evaluate them on the test set. Please refer to \ref{sec:app_proxy} for more analysis about proxy tasks and \ref{sec:full_mpii} for experiments on the full MPII~\citep{mpii} dataset.

\subsection{Evaluation of generalization capacity}\label{sec:exp_animalpose}

\textbf{Generalization to different domains.}
We investigate the generalization ability of our proposed model on domain transfer. To this end, we conduct experiments on AnimalPose~\citep{cao2019cross} dataset in Table~\ref{tab:animaltoanimal}. Specifically, we follow~\citep{cao2019cross} to choose one animal class (e.g. cat) as the target domain and the remaining four classes for the source domain. The images of the source domain are fully annotated, while the images of the target domain are unlabeled. 
For fair comparisons, we adopt the AlphaPose model~\citep{fang2017rmpe}, which uses ResNet-101~\citep{xiao2018simple} as the backbone. All models are trained with the pose-labeled human dataset involved. The AlphaPose baseline model is pre-trained on the human dataset and fine-tuned on the labeled source animal data. For pseudo-labeling based approaches~\citep{inoue2018cross,cao2019cross}, the unlabeled target animal data is used for pseudo-labeling. For domain adaptation approaches~\citep{tzeng2015simultaneous,long2016unsupervised}, the unlabeled target data is used for domain transfer. Please refer to~\cite{cao2019cross} for details about the compared approaches. We observe that our proposed approach consistently outperforms the previous state-of-the-art methods on cross-domain semi-supervised learning.

\begin{table}[htb]
    \caption{\textbf{Dataset 6: AnimalPose.} Evaluation of generalization capacity to the target unseen animal class. All results are obtained using the AlphaPose model~\citep{fang2017rmpe} with ResNet-101~\citep{he2016deep} as the backbone. Results marked with `*' are from~\cite{cao2019cross}.}
    \begin{center}
    \tabcolsep=2.0pt
    \begin{tabular}{@{}lccccc}
    \toprule
    & \multicolumn{5}{c}{mAP for each class}\\
    \cline{2-6}
    Method & cat & dog & sheep & cow & horse\\
    \hline
    AlphaPose Baseline$^{*}$~\citep{fang2017rmpe} & 37.6 & 37.3 &49.4 &50.3 &47.9 \\
    Dom Confusion$^{*}$~\citep{tzeng2015simultaneous} & 38.0 & 37.7 & 49.5 & 50.6 & 48.5\\ 
    Residual Transfer$^{*}$~\citep{long2016unsupervised} & 37.8 & 38.2 & 49.1 & 50.8 & 48.6\\
    CycleGAN+PL$^{*}$~\citep{inoue2018cross} & 35.9 & 36.7 & 48.0 & 50.1 & 48.1\\
    WS-CDA+PPLO$^{*}$~\citep{cao2019cross} & 42.3 & 41.0 & 54.7 & 57.3 & 53.1 \\
    PLACL (Ours) & \textbf{47.1} & \textbf{42.9} & \textbf{59.5} & \textbf{58.4} & \textbf{66.0} \\
    \bottomrule
    \end{tabular}
    \end{center}
    \label{tab:animaltoanimal}
\end{table}

\subsection{Analysis}

\begin{figure}[t]
	\centering
	\begin{subfigure}[b]{0.46\textwidth}
		\centering
		\includegraphics[width=0.99\textwidth]{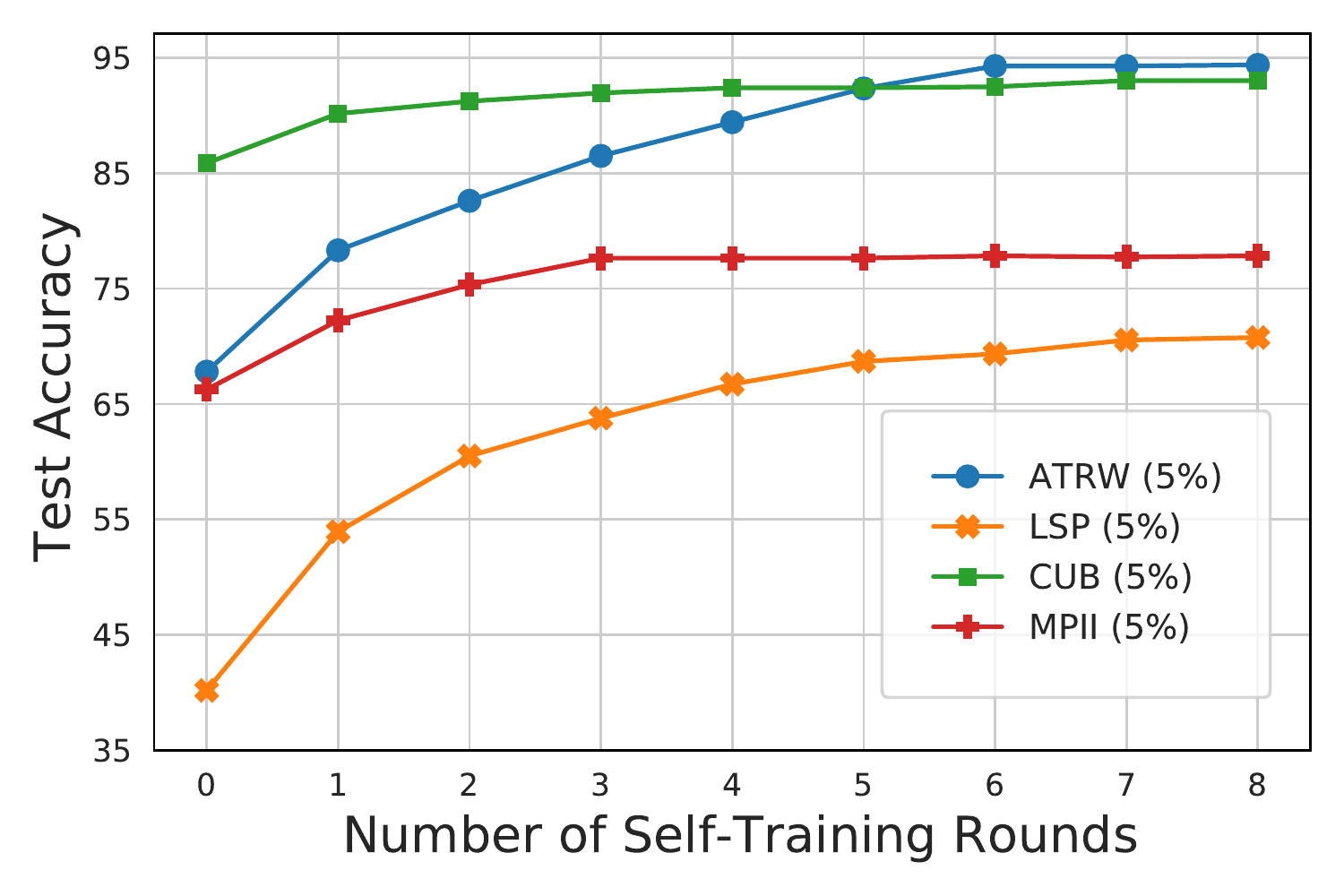}
		\caption{}
		\label{fig:round}
	\end{subfigure}
	\begin{subfigure}[b]{0.52\textwidth}
		\centering
		\includegraphics[width=0.99\textwidth]{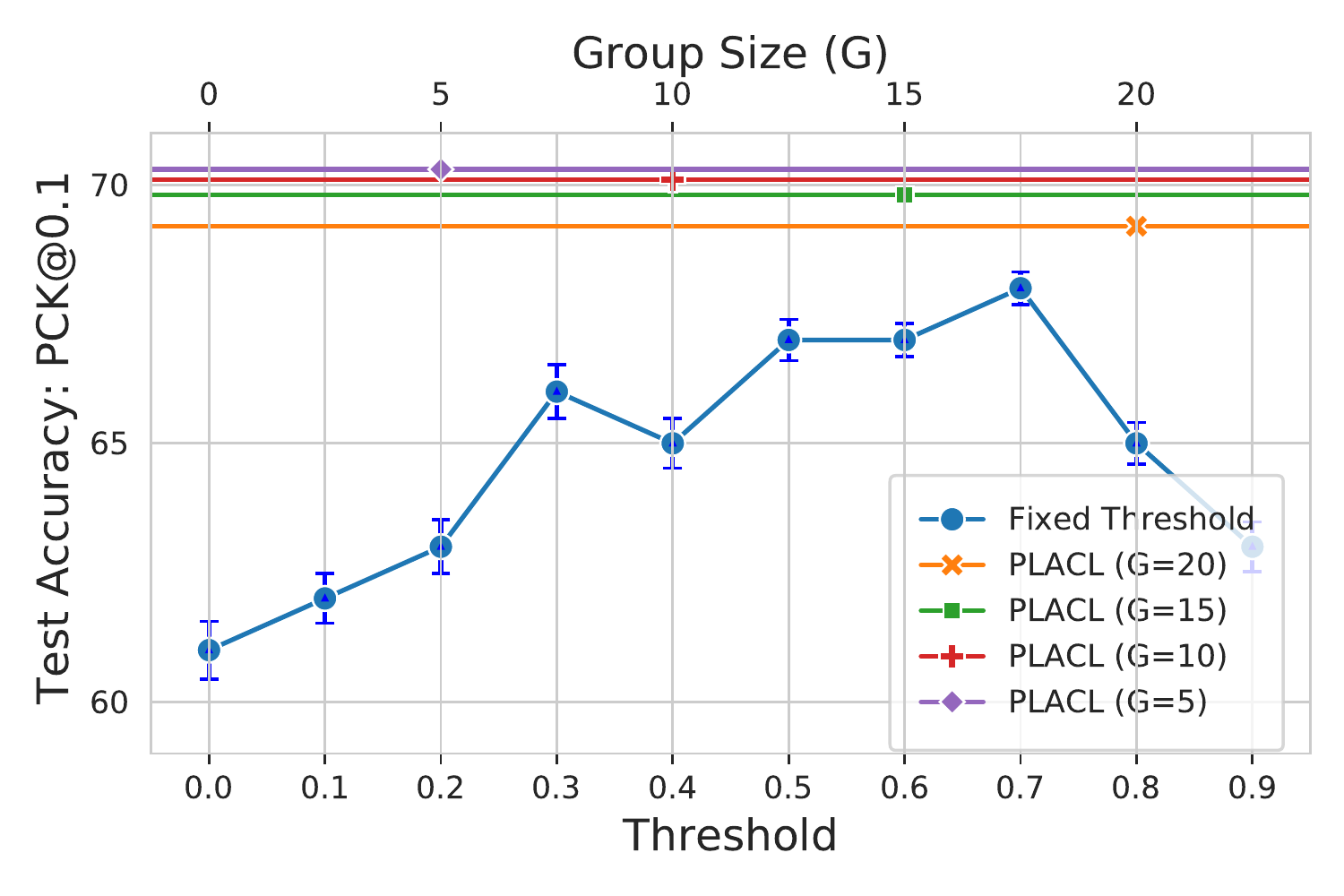}
		\caption{}
		\label{fig:threshold}
	\end{subfigure}
	\caption{(a) Accuracy on different datasets (5\% labeled data) with various rounds ($R$). (b) Comparisons with different fixed thresholds (blue dots) and different group sizes $G$. Experiments are conducted on LSPET dataset (5\% labeled data).}

\end{figure}

\textbf{Number of self-training rounds.} 
Along with the increasing of self-training rounds ($R$), the quality of the pseudo-labels gradually improves (see Fig.~\ref{fig:qualitative_results}) and the test accuracy increases (see Fig.~\ref{fig:round}) until saturation. The experiments are conducted on multiple datasets with 5\% labeled data. Interestingly, different datasets require a different number of rounds to achieve optimal, because four-legged animals (ATRW~\citep{ATRW}) have more pose variations than birds (CUB-200-2011~\citep{cub-200-2011}). We use $R=6$, because further increasing $R$ does not bring significant gains.

\textbf{Comparisons with pseudo-labeling with different fixed thresholds.} As shown in Fig.~\ref{fig:threshold}, we compare our PLACL with 10 static thresholds from $0.0$ to $0.9$. We observe that PLACL clearly outperforms all these fixed threshold alternatives. Moreover, we find that the accuracy will be significantly affected by different thresholds (61.01\% PCK for $\gamma=0.0$ vs 67.35\% PCK for $\gamma=0.7$).

\textbf{Choice of epoch group size}. In Fig.~\ref{fig:threshold}, we also compare the performance of different epoch group sizes ($G$). We empirically find that smaller $G$ will produce better performance, but at the cost of increased search space. We choose $G=10$ to trade-off between accuracy and efficiency.

\subsection{Ablation Studies}

In Table~\ref{tab:ablation}, we present ablation studies to measure the contribution of each component. All ablative experiments are conducted on LSPET~\citep{lsp-extended} dataset with 5\% labeled data. PCK@0.1 is adopted as the evaluation metric.

\textbf{Effect of cross-training strategy.} We find that without cross-training strategy the accuracy significantly drops from $70.76$ to $67.13$, due to noise accumulation over time. Especially, we find that there is little to no improvement after the first self-training round.

\textbf{Effect of curriculum learning.} We compare PLACL with the alternative that only searches for a fixed threshold via RL. We find that using dynamic thresholds improves upon the fixed-threshold alternative by a large margin, which validates the effectiveness of curriculum learning.

\textbf{Effect of parameter search.} We compare PPO2~\citep{schulman2017proximal} search with \textit{Random Search} (w/o PPO2 search). We randomly sampled $T \times M$ curricula and pick out the best one for comparisons. PPO2 search obtains much better performance ($70.76$ vs $65.42$). This indicates that the searching problem is non-trivial and that our searching algorithm is very effective.

We also compare with manually designed curricula whose thresholds are gradually decreasing on the epoch level. We tried five curricula with different decrease slopes and reported the best one as \textit{Manually Design}. We observe that PLACL significantly outperforms the manually-designed curriculum baseline which validates the effectiveness of automatic curriculum search.

\begin{table}[htb]
    \caption{Ablation studies on LSPET dataset with 5\% labeled data. }
    \begin{center}
    \tabcolsep=2.0pt
    \begin{tabular}{@{}lc}
    \toprule
    Method & PCK@0.1  \\
    \hline
    PLACL, w/o cross-training strategy & 67.13 \\
    PLACL, w/o curriculum learning & 68.51 \\
    PLACL, w/o PPO2 search (Random Search) & 65.42\\
    PLACL, w/o PPO2 search (Manually Design) & 65.71\\
    \hline
    PLACL, full method & 70.76 \\ 
    \bottomrule
    \end{tabular}
    \end{center}
    \label{tab:ablation}
\end{table}

\section{Conclusions}

We propose a novel Pseudo-labeled Auto-Curriculum Learning (PLACL) for the task of semi-supervised keypoint localization. We propose to learn a curriculum to automatically select reliable pseudo-labels and propose cross-training strategy to mitigate the confirmation bias problem.
Extensive experiments on 6 diverse datasets validate the effectiveness and versatility of the proposed method.
We believe that our proposed approach is generic and we plan to investigate the applicability of PLACL on other visual tasks, such as object detection and semantic segmentation.

~\\
\textbf{Acknowledgement.} We thank Lumin Xu and Wang Zeng for their valuable feedback to the paper. Ping Luo was supported by the Research Donation from SenseTime and the General Research Fund of HK No.27208720. Wanli Ouyang was supported by the Australian Research Council Grant DP200103223, FT210100228, and Australian Medical Research Future Fund MRFAI000085.

{\small
\bibliographystyle{iclr2022_conference}
\bibliography{iclr2022_conference}
}

\clearpage
\appendix
\section{Appendix}

\subsection{Pseudo-code for PLACL algorithm}
Here we present the pseudo-code for the proposed Pseudo-Labeled Auto-Curriculum Learning (PLACL) algorithm. In PLACL, the keypoint localization network and the curriculum policy are jointly optimized. 
In the inner-loop, we optimize the keypoint localization network as shown in Algorithm~\ref{alg:inner}. In the outer-loop, the curriculum policy is updated according to the performance of the keypoint localization network, as shown in Algorithm~\ref{alg:PLACL}.

\begin{center}
\scalebox{0.95}{
    \begin{minipage}{0.9\linewidth}
      \begin{algorithm}[H]
	\label{alg:inner}
	\KwIn{}
	\quad \quad 1. Labeled data $\mathbb{D}_{l}$, pseudo-labeled data $\tilde{\mathbb{D}}_{u}$. \\
	\quad \quad 2. Number of training groups per round $N_G$; \\
	\quad \quad 3. Curriculum $\Gamma$; \\
	\KwOut{Obtained optimal network weights $\omega^{*}(\Gamma)$.}
	\BlankLine
	Random initialization of network weights \;
	\For {$g=1$ \textnormal{\textbf{to}} $N_G$ }{
		Update current threshold $\gamma_g = \Gamma[g]$ \;
	    Compute data selection vector $\boldsymbol{g}$ (Eq.~\ref{eq:selection_criterion}) \;
	    Construct the training set $\mathbb{D}_{train}$ (Eq.~\ref{eq:dataset_train}) \;
	    Update network weights $\omega$ via back-propagation\;
	}
	\Return $\omega^{*}(\Gamma)$ \;
	\caption{Inner-loop network training}
\end{algorithm}
\end{minipage}
}
\end{center}

\begin{center}
\scalebox{0.95}{
    \begin{minipage}{0.9\linewidth}
      \begin{algorithm}[H]
	\label{alg:PLACL}
	\KwIn{}
	\quad \quad 1. Labeled data $\mathbb{D}_{l}$, unlabeled data $\mathbb{D}_{u}$. \\
	\quad \quad 2. Initialized distribution $\mu$ and $\sigma^2$. \\
	\quad \quad 3. Number of rounds $R$. \\
	\quad \quad 3. Searching steps $T$; Sampling number $M$. \\
	\quad \quad 4. Evaluation metric (e.g. PCK@0.1) $\xi$. \\
	\KwOut{Obtained the optimal curriculum $\Gamma^{*}$ and the final network $\Theta_{\omega^*}$.}
	\BlankLine
	Initialize $(\Gamma^{0})^*$ with all zeros. \\
	Pre-train keypoint localization network $\Theta_{\omega}^{0}$ using labeled data $\mathbb{D}_{l}$. \\
    \For {$r=1$ \textnormal{\textbf{to}} $R$}{
        \If { $r\%2 == 1$}{
		    Predict keypoint pseudo-labels $\tilde{\mathbb{D}}_{u} ^{(1)}$ with $\Theta_{\omega^*}^{r-1}$\;
            $\tilde{\mathbb{D}}_{u} = \tilde{\mathbb{D}}_{u}^{(1)}$
        }
        \Else {
		    Predict keypoint pseudo-labels $\tilde{\mathbb{D}}_{u} ^{(2)}$ with $\Theta_{\omega^*}^{r-1}$\;
            $\tilde{\mathbb{D}}_{u} = \tilde{\mathbb{D}}_{u}^{(2)}$
        }
        
    	\For {$t=1$ \textnormal{\textbf{to}} T }{
    		\For {$j=1$ \textnormal{\textbf{to}} M}{
    			Sample parameter $\Delta \Gamma_{j, t}^{r} \sim \mathcal{N}_{\text {trunc }[0,1]}\left(\mu_{t}^{r}, \sigma^{2} I\right)$ \;
    			$\Gamma^{r}_{j, t} = \Delta \Gamma^{r}_{j, t} + (\Gamma^{r-1})^*$ \;
			    Get $\Theta_{\omega^*, j, t}^{r}$ via inner-loop network training using $\Gamma^{r}_{j, t}$ (Alg.~\ref{alg:inner}) \;
    			Compute the evaluation metric $\xi\left(\Gamma_{j, t}^r\right) = \xi(\Theta_{\omega^*, j, t}^{r}; \mathbb{D}_\text{val})$ \;
    		}
    		Compute the objective function $J\left(\mu\right)$ (Eq.~\ref{eq:objective-function})\;
    		Update $\mu_{t+1}^r \leftarrow \mu_t^r +\alpha \nabla_{\mu^r} J\left(\mu^r\right)$ \;
    	}
    	$(\Gamma^{r})^*=\arg \max _{\mu_{t}^r} \sum_{j=1}^{M} \xi\left(\Gamma_{j, t}^r \right), \forall t=1, \ldots, T$ \;
    	Get $\Theta_{\omega^*}^{r}$ via network training using $(\Gamma^{r})^*$ (Alg.~\ref{alg:inner}) \;
	}
	\Return $\Gamma^{*}= [(\Gamma^{1})^*, \dots, (\Gamma^{R})^*]$ and $\Theta_{\omega^*}^{R}$ \;
	\caption{Pseudo-Labeled Auto-Curriculum Learning (PLACL)}
\end{algorithm}
\end{minipage}
}
\end{center}

\subsection{Visualization of Pseudo-Labeled Samples}

In order to provide a better illustration of how pseudo-labels evolve in self-training rounds, we visualize some pseudo-labeled samples for different datasets. We observe that the quality of pseudo-labels gradually improves with the increase of the self-training rounds.

\begin{figure}[h]
    \begin{center}
    \includegraphics[width=.98\linewidth]{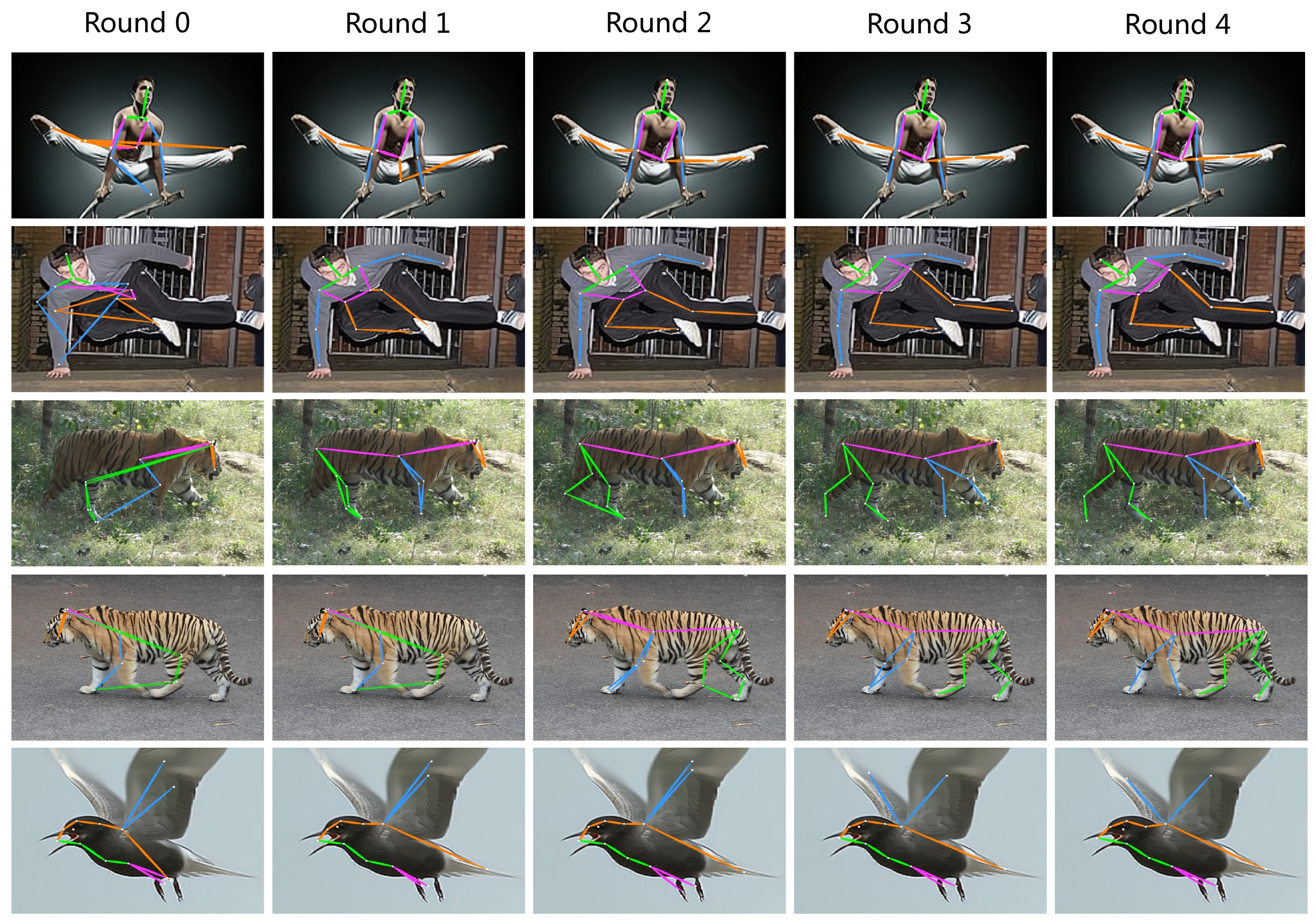}
    \end{center}
    \caption{Visualization of how pseudo-labels evolve in self-training rounds.} 
    \label{fig:qualitative_results}
\end{figure}

\subsection{Generalization to different keypoint localization models.}

Table~\ref{tab:model} shows the improvement when PLACL is applied to the recent state-of-the-art keypoint localization models which vary in model architectures and training/testing techniques. The experiments are conducted on LSPET dataset with 5\% labeled images and 95\% unlabeled images.
We show that PLACL consistently improves the performance of the state-of-the-art approaches by a large margin. PLACL does not require any knowledge of the keypoint localization models, making it easy to use in practice.

\begin{table}[htb]
    \caption{Performance improvement of different keypoint localization methods by PLACL. Experiments are conducted on CUB-200-2011 dataset (5\% labeled data) with PCK@0.1 as the metric.}
    \begin{center}
    \tabcolsep=2.0pt
    \begin{tabular}{@{}lccc}
    \toprule
    Method & Backbone & w/o PLACL & w/ PLACL \\
    \hline
    SimpleBaseline~\citep{xiao2018simple} & ResNet-50 &    79.16        & 93.51 \\
    SimpleBaseline~\citep{xiao2018simple} & ResNet-101 & 81.34 & 93.66  \\
    SimpleBaseline~\citep{xiao2018simple} & ResNet-152 & 86.15 & 94.27 \\
    HRNet~\citep{sun2019deep} & HRNet-w32 & 85.86 & 93.01 \\
    HRNet~\citep{sun2019deep} & HRNet-w48 & 85.89 & 94.26 \\
    DARK~\citep{zhang2020distribution} & HRNet-w32 & 86.67 & 94.18\\
    \bottomrule
    \end{tabular}
    \end{center}
    \label{tab:model}
\end{table}

\subsection{Analysis of proxy tasks}
\label{sec:app_proxy}

There are a lot of methods that target at improving the searching efficiency in literature, \eg using reduced-training proxy tasks (input size, model size, training samples, and training epochs can be reduced~\citep{zhou2020econas}). 
With these techniques, we are able to get orders of magnitude less computation cost, but can still match the performance. Therefore, we believe that the scalability is not a problem. For example, in order to further reduce the complexity, we use a light proxy task (with reduced training samples) for the RL search process. Specifically, we randomly select a small proportion of the training data (\eg 5k images) for efficient curriculum search. After the search procedure, we re-train the keypoint networks with the searched curriculum on the full training set and evaluate them on the test set. 

As shown in Table~\ref{tab:proxy}, we randomly select different number of training images (5k and 10k) for RL curriculum search. We find that reducing the number of training images by half (from 10k to 5k) does not decrease the final performance much, which validates the effect of using proxy tasks. 
Such a strategy enables us to apply the proposed PLACL approach to large-scale datasets, such as MS-COCO~\citep{coco} and the full MPII~\citep{mpii} datasets.

\begin{table*}[h]
\caption{We randomly select different number of training images (\#Images) for RL curriculum search, and re-train the keypoint networks with the searched curriculum on the full training set.}
\label{tab:proxy}
\begin{center} 
\begin{tabular}{llllll} 
\#Images & \multicolumn{1}{c}{5\%} & \multicolumn{1}{c}{10\%} & \multicolumn{1}{c}{20\%} & \multicolumn{1}{c}{50\%} & \multicolumn{1}{c}{100\%} \\ 

\toprule 
\multicolumn{6}{c}{\textbf{LSPET } } \\ 
5k    & 70.72$\pm$1.49 & \textbf{71.93$\pm$1.17} & 72.24$\pm$0.82 & 72.71$\pm$1.19 & - \\ 
10k    & \textbf{70.76$\pm$1.47}  & 71.91$\pm$1.15  & \textbf{72.30$\pm$0.88}  & \textbf{72.73$\pm$1.23}  & - \\

\toprule 
\multicolumn{6}{c}{\textbf{MS-COCO'2017}} \\ 
5k    & \textbf{69.39$\pm$1.03}  & 70.11$\pm$0.89  & \textbf{71.84$\pm$0.66}  & 73.42$\pm$0.57  & - \\ 
10k     & 69.24$\pm$1.02 & \textbf{70.12$\pm$0.87}  & 71.61$\pm$0.63 & \textbf{73.43$\pm$0.61} & - \\

\bottomrule
\end{tabular} 
\end{center}
\end{table*}

\subsection{Experiments on the full MPII dataset}
\label{sec:full_mpii}
As shown in Table~\ref{tab:full_mpii}, we provide the results on the full MPII~\citep{mpii} dataset. 
Since the codes of SSKL~\citep{moskvyak2020semi} are not publicly available, we only compare with CL~\citep{cascante2020curriculum} in the experiments. We see that our proposed PLACL consistently outperforms CL, especially for low labeled data regime (5\% and 10\%). Note that our results on the full MPII are obtained using reduced-training proxy tasks, \ie we use 5K images for RL curriculum search, and re-train the model with the obtained curriculum on the full training set.

\begin{table*}[h]
\caption{Comparisons with CL on the full MPII dataset.}
\label{tab:full_mpii}
\begin{center} 
\resizebox{1.\textwidth}{!}{
\begin{tabular}{llllll} 
 Method & \multicolumn{1}{c}{5\%} & \multicolumn{1}{c}{10\%} & \multicolumn{1}{c}{20\%} & \multicolumn{1}{c}{50\%} & \multicolumn{1}{c}{100\%} \\ 
\toprule 
\multicolumn{6}{c}{\textbf{Full MPII} } \\ 
HRNet~\citep{sun2019deep}   & 78.00$\pm$1.35  & 81.89$\pm$0.94 & 82.94$\pm$0.67 & 88.34$\pm$0.45 & 89.76$\pm$0.17    \\ 
CL~\citep{cascante2020curriculum}   & 80.38$\pm$1.31 & 83.06$\pm$0.89 & 84.57$\pm$0.68 & 88.72$\pm$0.34 & \\ 
PLACL (Ours) & \textbf{82.21$\pm$1.22}  & \textbf{85.42$\pm$0.85}  & \textbf{86.24$\pm$0.56}  & \textbf{89.16$\pm$0.21}  & - \\

\bottomrule
\end{tabular} 
}
\end{center}
\end{table*}

\end{document}